\documentclass[conference]{IEEEtran}
\IEEEoverridecommandlockouts
\usepackage{cite}
\usepackage{amsmath,amssymb,amsfonts}
\usepackage{algorithmic}
\usepackage{graphicx}
\usepackage{amssymb}
\usepackage{pifont}
\newcommand{\xmark}{\ding{55}}%
\usepackage{textcomp}
\usepackage{mathtools}
\DeclarePairedDelimiter\ceil{\lceil}{\rceil}
\usepackage{pbox}
\usepackage{makecell}
\usepackage{textcomp}
\usepackage{xcolor}
\def\BibTeX{{\rm B\kern-.05em{\sc i\kern-.025em b}\kern-.08em
    T\kern-.1667em\lower.7ex\hbox{E}\kern-.125emX}}
\newtheorem{theorem}{Theorem}
\begin{document}

\title{K-nearest Multi-agent Deep Reinforcement Learning  for Collaborative Tasks with a Variable Number of Agents
}

\author{\IEEEauthorblockN{ Hamed Khorasgani}
\IEEEauthorblockA{\textit{Hitachi Industrial AI Lab} \\
Santa Clara, CA, USA }
\and
\IEEEauthorblockN{ Haiyan Wang}
\IEEEauthorblockA{\textit{Hitachi Industrial AI Lab} \\
Santa Clara, CA, USA }
\and 
\IEEEauthorblockN{ Hsiu-Khuern Tang}
\IEEEauthorblockA{\textit{Hitachi Industrial AI Lab} \\
Santa Clara, CA, USA }
\and 
\IEEEauthorblockN{     Chetan Gupta}
\IEEEauthorblockA{\textit{Hitachi Industrial AI Lab} \\
 Santa Clara, CA, USA 
}}

\maketitle

\begin{abstract}
Traditionally, the performance of multi-agent deep reinforcement learning algorithms are   demonstrated and validated in gaming  environments where we often  have a fixed number of agents. 
In many industrial applications, the number of available agents can change at any given day and even when the number of agents is known ahead of time, it is common for an agent to break during the operation and become unavailable for a period of time.
In this paper, we propose a new deep reinforcement learning algorithm for multi-agent  collaborative tasks with a variable number of agents.  We demonstrate the application of our algorithm using a fleet management simulator developed by Hitachi to generate realistic scenarios in a production site.  
\end{abstract}


\section{Introduction}
Multi-agent deep reinforcement (RL) 
algorithms have shown superhuman performance in gaming  environments such as the Dota 2 video
game \cite{Dota} and StarCraft II \cite{StarCraft}. For  industrial applications, millions of dollars can be saved by making small
improvements in productivity. The unprecedented performance of multi-agent deep RL in learning
sophisticated policies in collaborative and competitive games  makes one wonder why  industries are reluctant in using multi-agent deep RL algorithms to achieve more efficient policies. Recently, Dulac-Arnold, Mankowitz, and Hester listed nine
main challenges in applying deep RL in real-life applications  \cite{Challenges}. Dulac-Arnold et al.  formulated these challenges in a more  formal way  using a Markov Decision Process framework and
presented some existing work in the literature to solve them  \cite{Challenges2}. 
In previous work, we  discussed  
additional challenges in using deep RL in an industrial environment \cite{Challenges3}. Among them, multi-agent systems with a variable number of agents have received very limited attention from the research community \cite{variable quantity of agents}. 
However, agent failures are very common in industrial applications. An agent's failure during the operation or its return to the operation after maintenance or repair  can
change the number of operating agents abruptly. 

Moreover, a change in the production target can increase or decrease the number of operating agents in an industrial setting from one shift to the next. Therefore, it is critical to address the   systems with varying number of agents to make the application of multi-agent deep RL in the industrial setting a viable solution. 
In this paper, we propose a new deep RL multi-agent algorithm which can address the challenge of multi-agent systems with a variable number of agents. Moreover, our solution has the capacity to learn an optimal policy when we have action dependency among neighboring agents.   This can be very helpful for  industrial settings with collaborating agents because  it is often impossible for an agent to make optimal decisions  independent of the  agents with which  it has a  high level of interaction.

 Section \ref{Multi-agent Deep RL Literature review} reviews deep RL solutions for multi-agent systems. In this section we present the pros and cons of common multi-agent deep RL algorithms and justify the need for developing our proposed solution. Section \ref{K-nearest Multi-agent  RL} presents the  K-nearest multi-agent deep RL    as a new algorithm to address some of the challenges which have prevented  the  industrial applications to fully benefit from deep RL algorithms. Section \ref{Case Study} presents the application of  K-nearest multi-agent RL in a fleet traffic management  system.   The fleet traffic control problem is   a real industrial challenge in a Hitachi production site  which motivated us to develop the K-nearest multi-agent RL algorithm. Section \ref{Conclusions} concludes the paper.

\section{Multi-agent Deep RL Literature review }
\label{Multi-agent Deep RL Literature review}

\begin{table*}[htbp]
\caption{Comparison of different multi-agent RL solutions for industrial applications }
\begin{center}
\begin{tabular}{|c|c|c|c|c|}
\hline
\textbf{RL}&\multicolumn{4}{|c|}{\textbf{Challenge}} \\
\cline{2-5} 
\textbf{Methods} & \textbf{\textit{Variable number of agents}}& \textbf{\textit{Non-stationary training}}& \textbf{\textit{Action dependency}} & \textbf{\textit{Model management}}
 \\
\hline
Centralized  & \xmark& $ \checkmark $  & $ \checkmark $  &$ \checkmark $  \\
\hline
Centralized training/decentralized execution & \xmark&  $ \checkmark $  &\xmark   &$ \checkmark $  \\
\hline
Independent Q-learning (IDQ) &$ \checkmark $  &  \xmark   & \xmark   & \xmark   \\
\hline
 IDQ + weight-sharing &$ \checkmark $ & $ \checkmark $  & \xmark  &$ \checkmark $  \\
\hline
K-nearest RL + weight-sharing & $ \checkmark $  & $ \checkmark $  & $ \checkmark $  &$ \checkmark $  \\
\hline
\multicolumn{4}{l}{  \pbox{20cm}{ $\checkmark $: the method  addresses the challenge. \\ 
 \xmark: the method cannot address the challenge.}}
\end{tabular}
\label{tab1}
\end{center}
\end{table*}
Many real-world problems are  multi-agent problems. When the number of agents is small, it
is possible to model multi-agent problems using a centralized approach wherein we train a centralized
policy over the agents’ joint observations and produce  a joint set of actions.
In the centralized approaches, the policy network generates all of the agents' actions using the global observations. Therefore, these  methods  can achieve optimal policies even when the agents' actions are interdependent. 

However, one can imagine that
 the centralized approaches  do not scale well for many applications and we will quickly have state and action
spaces with very large dimensions. Moreover, they are not practical solutions to address industrial problems, in which  the number of agents is not
constant.
In many real-world applications, the number of  agents can change  based on the  product's demand, the equipment's scheduled maintenance and repairs, unexpected breakdowns, operator  availability,  etc. 

The centralized RL algorithms consider the global feature space as the union of the agents' features and generate actions for all of the agents simultaneously. Therefore, these methods assume that the number of agents is fixed. Removing an agent leaves a hole
in the network. Moreover, we cannot add additional agents during an episode, which can be the
case in many industries. A trivial solution to this problem is to learn different policy networks
for each different number of agents, so that we can enact the appropriate policy when the number of agents
changes. However, the range of agents could  vary  widely in an industrial application.  To have a model
trained and updated for each combination is expensive or even infeasible,  especially for a large system 
with hundreds of agents.

A more realistic approach is to use an autonomous learner for
each agent such as independent deep-Q-network (IDQN) which distinguishes agents by identities \cite{IDQ}. Even though the
independent learners address the challenges of large  action
space  and  a variable number of agents  to some extent, they suffer from the convergence
point of view as the environments become non-stationary. In fact, these algorithms model the other
agents as part of the environment and, therefore, the policy networks must  chase moving targets as
the agents’ behaviors change during the training \cite{tsitsiklis1994asynchronous}.

Moreover, when we  use the IDQN framework, we force the  agents to take actions independently.
Therefore, unlike  the centralized RL algorithms,   IDQN cannot achieve an optimal policy when an optimal action for an agent depends on the other agents' actions.   This limits the performance of IDQN in many collaborative industrial applications wherein each agent is interacting  with  several other   agents' and must consider their  behavior  in its decision-making  to maximize the overall performance.  Finally, from a practical point of view, it  can be difficult  to train and maintain a separate policy network for each agent,  especially in large-scale industrial systems in which there  can be  hundreds of agents.

To address the convergence problem, centralized learning with decentralized execution approaches have been proposed in recent years. In these
methods, a centralized learning approach is combined with a decentralized execution mechanism
to have the best of both worlds. Lowe et al.  proposed a multi-agent deep deterministic policy
gradient (MADDPG), which includes a centralized critic network and decentralized actor networks
for the agents \cite{Multi-agent actor critic}. Sunehag et al.  proposed a linear additive value decomposition approach wherein 
the total Q value is modeled as a sum of individual agents’ Q values \cite{Value-decomposition}. Rashid et al.  proposed the 
Q-MIX network, which allows for a richer mixing of Q agents compared to the linear additive value decomposition \cite{Qmix}. 

Even though the centralized learning with decentralized execution approaches have shown promising results in many applications, they have the same limitations as the centralized RL methods  when it comes to the industrial problems in which  the number of agents is not
constant. Moreover, because the execution is decentralized, they  cannot solve multi-agent problems  in which the optimal action for an agent depends upon  the other agents' actions.

Foerster, Assael, Freitas, and Whiteson  proposed a single network with shared parameters to reduce the number of
learned parameters and speed up the learning \cite{Foerster}. In a previous work, we applied a similar approach to address a dynamic dispatching problem in an industrial multi-agent environment  \cite{HitachiDynamicDispatching}. Having a shared policy among agents solves the agent
failure challenge and addresses  the  non-stationary problem to some extent. To further reduce  the non-stationary
problem, Foerster, Assael, Freitas, and Whiteson disabled experience replay \cite{Foerster}. A single network with shared parameters also solves the model management challenge because it only requires training and maintaining one model for all of the agents.  However, like the autonomous learner algorithms and  the  centralized learning with decentralized execution approaches, this approach cannot achieve optimal policy when optimal action for an agent depends upon the other agents' actions. Table \ref{tab1} presents a summary of  advantages and disadvantages  of different multi-agent RL solutions. 

\section{K-nearest Multi-agent  RL}
\label{K-nearest Multi-agent  RL}
\begin{figure*}[htbp]
\includegraphics[scale=.45]{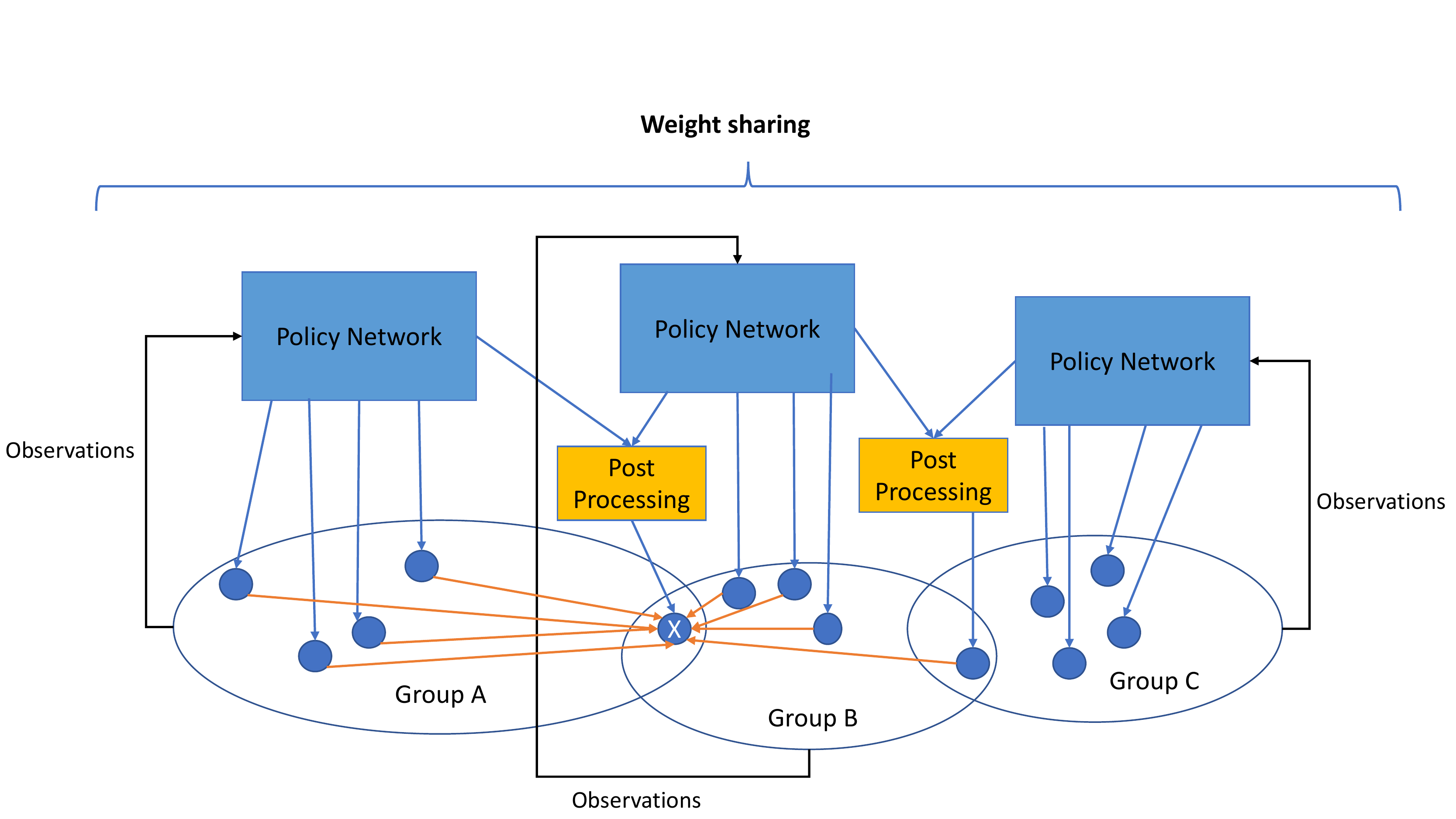}
\centering
\caption{K-nearest Multi-agent deep RL. When an agents belongs to more than one group, the post processing unit derives the final action  using the  weighted average of the actions proposed by different groups. The actions are weighted inversely proportional to the agent's distance from each group's agents. The closer the agent is to the group, the group's action gets a higher weight.  }
\label{fig1}
\end{figure*}
In this section, we present K-nearest deep RL which is designed to address two main challenges in multi-agent industrial problems: 1) varying  number of agents and  2) action dependency among the agents. A varying number of agents is one of the main challenges that industries face when dealing with multi-agent problems. This problem often arises because of a change in the production plan or equipment failures. Action dependency is another major issue in many industrial applications. This occurs when the optimal action of an agent depends upon the action another agent takes. On one side of the spectrum, independent learner methods can address the variable agent challenge, but they fail to perform well when there is action dependency among the agents. On  another side of the spectrum, centralized RL algorithms can fully address the action dependency. However, they are not suitable for situations in which  there are a  variable number of agents. K-nearest RL proposes a trade-off solution which combines the best of both worlds.

We observed that in many industrial applications with collaborative agents the optimal  actions an agent takes are  highly dependent upon the nearby agents’ actions,  but they are not dependent upon  the actions taken by other agents that the agent is not interacting with. For example, in the traffic control problem, the optimal speed for a given vehicle can dramatically change if the speed of a nearby vehicle changes but the  speed of two vehicles on different  sides of the production site hardly have   any impact on one another. The K-nearest RL aims to use this locality to address action dependency when it matters while avoiding challenges that come with the centralized  solutions. 
\subsection{K-nearest agents}
Our proposed algorithm  finds the k-nearest agents to each agent and generates their actions using a common policy network. 
The metric with which to  measure the distance between two agents depends upon  the problem. For example, in a traffic control problem where each vehicle  is an agent, the distance between two vehicles can be defined based upon the traveling distance between them, but in a production line in which each operating machine is an agent, the distance between each two agents  may be defined as the time it takes for a  product to travel between the machines. 
Consider a system with $N$ number of agents. After grouping each agent with its k-nearest agents, we will have $m$ unique groups, wherein   the number of agents in each group is equal  to $k+1$. Figure \ref{fig1} shows an example wherein  $N=13$, $k=4$   and at this time step, we have three groups; $m=3$.

The number of groups, $m$, would be different based upon the agents' formation. The following theorem provides the upper bound and lower bound for $m$.  

\begin{theorem}
\label{theorem}
At  any given time, the number of groups, $m$, in k-nearest RL algorithm is:
\begin{equation}
\ceil{\frac{N}{(k+ 1)}} \le m \le N, 
\label{m_eq}
\end{equation}
where $\ceil{x}$ represents the ceil function of $x$ which is  the smallest integer that is greater than or equal to $x$. 
\end{theorem}
\textit{Proof:} see Appendix A.

\subsection{Weight sharing}
Since all the groups have the same number of agents equal to $k+1$, we can have a policy network with $k+1$ outputs to assign actions of agents in each group. 
Similar to  \cite{Foerster}, different groups have the same policy networks. The weight sharing makes our model easy to train and maintain. Moreover, our solution can address dependent actions because one network selects  optimal actions for each agent and the nearby agents.  
Note that independent learner solutions are special cases of K-nearest RL when $k=0$, and the centralized RL algorithms are the special case of K-nearest RL when $k=N-1$.

\subsection{Post processing}
In our algorithm, it is possible that an agent belongs to more than one group. In this case, the agent may receive different commands from policy networks of different groups that it belongs to. We can address this problem using different aggregation strategies  based on the application. In the traffic control case study, we use the weighted average of different actions. The weight of each group is inversely proportional to the total distance of the agent from the other agents in  the group.  Therefore, the agent gives a higher weight to the group with which it is closer. Figure \ref{fig1} shows two examples of an agent belonging to more than one group. Consider agent x. At this time step, x belongs to group A and group B. Using our weighted sum aggregation method, agent x gives higher weight to the action proposed by policy network of group B because it is located closer to the agents of this  group  than the agents of group A. 
The focus of this work is on continuous action space.  However, it is possible to develop an aggregation method for discrete action space as well. For example, we can simply choose the action from the group that the agent is closest to.

\subsection{Off-policy training}
The K-nearest RL algorithm learns a policy for each group of k agents. Using an off-policy RL algorithm to train the network, we must  save the observation of k agents, the actions of the k agents, the next observations of the k agents and the total reward of the k agents in the replay buffer. We consider the reward of each group equal  to the total reward of the k agents to encourage them to collaborate with other local agents in the group in order to maximize the common profit. 
We keep track of the group observation and reward  in the next step and save the results in the replay buffer even though  after the agents take the actions,  distances between different agents  may change and  in the next step we may have a different number of groups. All of the group policy networks share the samples and the weights during the  off-policy training. The idea is to learn a policy that generates an optimal solution for any $k+1$ nearby  agents. 

We use  the Soft Actor Critic (SAC) algorithm  \cite{SAC} which is an  off-policy deep RL algorithm that uses  maximum entropy to improve robustness as our RL algorithm to learn the network policy for each group. The SAC algorithm   achieves state-of-the-art performance on various  continuous action space benchmarks. Moreover, it   is very stable, and its performance does not change significantly     across   different training episodes. Nonetheless, our solution is  not  dependent upon  the SAC algorithm and it can be used with any off-policy RL algorithm. 

\subsection{The overall algorithm}
Here are  the  steps to train the  K-nearest RL  algorithm:
\begin{enumerate}
    \item Identify the k-nearest agents to each agent. For a system with $N$ agents, this will generate $N$ groups wherein  each group has $k+1$ agents. The distance  can be defined based on agent interactions for any given problem. For example, in the traffic control, the traveling  distance between two vehicles can represent the distance between them. 
    \item Remove redundant groups. Two groups are redundant if they have the same agents. This will reduce the number of groups to $m$ (see equation (\ref{m_eq})).
    \item Train an  SAC  policy network with k+1 actions to control each group. We use weight sharing for each group. The reward for each group is the total rewards of its agents. This encourages collaboration among the agents.
    \item Apply actions generated by each policy network to the agents; if an agent belongs  to more than one group, apply a weighted action. The weights are based on how close the agent is to each group's agents. 
    \item Go back to the first step.
\end{enumerate}

The k-nearest RL can be applied to any multi-agent problem in which 
we can define the nearby agents as a subset of highly interacting agents. An  example for these systems is a fleet traffic control problem where nearby vehicles have much higher impact on each other. In the next section, we demonstrate this application.

\section{Case Study}
\label{Case Study}
We developed the K-nearest multi-agent deep RL algorithm to address traffic control in a Hitachi fleet management system.  In this setting, different vehicles travel in a production site to transfer personnel and materials. The destination of each vehicle is set by a separate dispatching algorithm. Our goal is to set the speed for each vehicle in order to minimize the  traffic congestion and therefore,   maximize  the total number of cycles in each shift. Reducing the traffic congestion can lead to millions of dollars in cost  saving and reduce a significant amount of carbon emission.


Considering  the actuator delays which are the time  between when a vehicle receives  a speed command and the time that the  vehicle reaches  that  desired speed,   different roads and intersections in  different parts of the facility, and the multi-agent nature of the problem where each vehicle represents an agent, it is not feasible to solve this problem analytically. Multi-agent deep RL seems like a natural solution for  this problem. 
Therefore,  we developed the K-nearest multi-agent deep RL algorithm to utilize the potentials of  multi-agent deep RL while  addressing  the following challenges  in the system:
\begin{itemize}
    \item The number of available vehicles could change in any day and it is very expensive to have a different model for different combinations of vehicles  as each model has to be maintained over time.
    \item The actions of each vehicle (the speed of vehicle) cannot be set independent of the nearby vehicles because  an independent decision-making process can lead to a sub-optimal solution for the entire  system.  
\end{itemize}

The current default algorithm sets the speed of each vehicle equal to the maximum speed limit of  each road. Therefore, we call the default algorithm the maximum speed algorithm. Coming up with an algorithm to achieve a higher number of cycles per shift  than the maximum speed algorithm is challenging because the vehicles are not allowed to travel faster than the speed limit. In fact, our algorithm should increase the overall number of cycles by reducing some vehicles' speeds which may sound counter-intuitive. However, this is possible because higher speeds may lead to queuing and congestion further down the road. 

\begin{figure}[htbp]
\includegraphics[scale=.45]{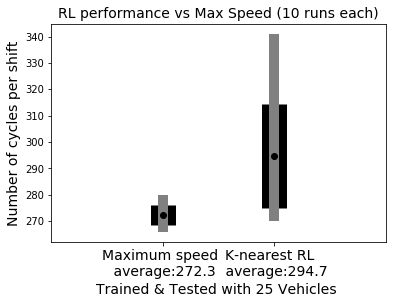}
\centering
\caption{Number of cycles in a shift for the baseline approach and our proposed k-nearest RL solution. The production facility had $25$ vehicles  during the training and the test. The grey lines show the maximum and minimum number of cycles for each method. The black lines show one standard deviation of cycles for each method. The black dots show the average of cycles for each method. The   k-nearest RL achieves much better performance than the baseline solution. This can lead to millions of dollars in cost  saving and significant reduction in the site's greenhouse gas emissions over years. }
\label{result_1}
\end{figure}

We use a simulator developed by Hitachi which is the digital twin of the actual production facility to train and test the K-nearest deep RL algorithm. The simulator is designed to mirror the  actual vehicles' speed profiles and the facility map. We set $k=4$ which means each group includes five vehicles. This hyper parameter is set based upon our understanding of the level of vehicle interactions in the facility  and the fact that the facility always has at least five traveling vehicles. For other problems, based upon the number of interacting agents and the minimum number of operating agents, we may need a higher or a lower $k$.

Each episode during the training represents a $12$-hour shift in the factory. We use the  SAC deep RL algorithm to learn the policy network for the groups. The observation for each group includes the location of each vehicle in the group, the distance between each pair of vehicles in the group, the current speed of each vehicle in the group, the time elapsed since each vehicle in the group has left the previous stop which could be a  loading site or a discharge site, the distance it traveled since  the last stop, and the distance it has left to travel to reach  its current goal determined by the dispatching algorithm.

We renew  the speed targets  every $60$ seconds. We keep track of vehicles in each group until the next step to  save the next observation.  
 The reward for each group is the total distance traveled by the vehicles in the group from the time  we set the speeds until the next time step. 
 \begin{equation}
r_j = \sum_{i=1} ^ {k+1} \delta_{ij},
\label{r_eq}
\end{equation}
where $r_j$ is the reward for group number $j$ and  $\delta_{ij}$ represents the distance traveled by  vehicle number $i$ in group   $j$ over the last $60$ seconds. We set the reward equal to the total distance traveled by  the group's  vehicles since the previous time step     for each group to encourage collaboration among the vehicles.

Fig. \ref{result_1} shows the number of cycles we achieve in one shift using the default maximum speed algorithm and our proposed K-nearest deep RL algorithm. The black line shows one standard deviation of each method over 10 runs. The gray line shows the minimum and maximum number of cycles for the 10 runs. We run each algorithm multiple times because there are several stochastic variables in the simulator,  such as the time it takes for the personnel or load to load or discharge which can affect the number of cycles. Fig. \ref{result_1} shows that  the K-nearest RL algorithm clearly outperforms the maximum speed algorithm. This is significant because the maximum speed is a very strong baseline method which most drivers choose intuitively.

\begin{figure}[htbp]
\includegraphics[scale=.45]{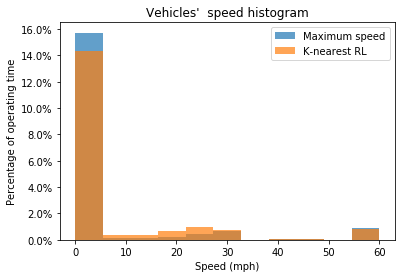}
\centering
\caption{Distribution of the operating speed for  $25$ vehicles during two consecutive 12-hours shifts. Compared to the K-nearest RL policy, the maximum speed policy  leads to higher traffic congestion and therefore, higher percentage of operating time  with less than 5 mph speed. }
\label{speedhist}
\end{figure}

Fig. \ref{speedhist} shows the histogram distribution of vehicles' speeds operating under the maximum speed algorithm and  the K-nearest RL policy for two consecutive 12-hours   shifts. When we  use the K-nearest RL policy, there are fewer moments at  which the vehicles are moving at the highest speed limit (60 mph). However, there are also much fewer instances that vehicles are stopped or moving  less than 5 mph due to congestion. This figure shows how K-nearest RL improves the overall performance by occasionally lowering the vehicles' speed below the maximum speed limit. 

\begin{figure}[htbp]
\includegraphics[scale=.45]{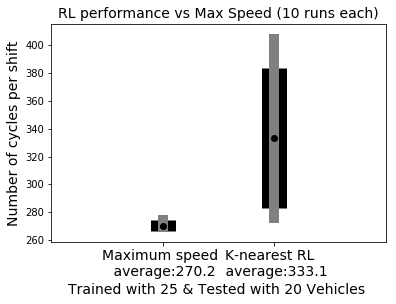}
\centering
\caption{Number of cycles in a shift for the baseline approach and our proposed k-nearest RL solution when  the production facility had $25$ vehicles  during the training and 20 vehicles during the test. The results show the k-nearest deep RL algorithm can perform well in an environment with variable number of agents. Even though the k-nearest RL algorithm is not trained with  20 agents, it still outperforms the very strong baseline method.  }
\label{result_2}
\end{figure}
The simulator operated with $25$ vehicles during the training. To demonstrate  the robustness of K-nearest RL, we run another experiment in which we test the trained policy with $5$ fewer vehicles.   Fig. \ref{result_2} shows  that the K-nearest policy  trained with $25$ vehicles outperforms the maximum speed policy even when   the number of vehicles is reduced  to $20$. This demonstrate the robustness of our algorithm and the fact that it can be applied in environments with a variable number of agents.

\section{Conclusions}
\label{Conclusions}
In this work, we presented K-nearest deep RL to address challenges of a variable number of agents, action dependency, model management, and non-stationary training in order to make deep multi-agent RL more applicable in the industrial settings. First, K-nearest deep RL can perform in an environment with  a variable number of agents as long as the total number of agents does not fall below the  number of agents in each group, $k+1$. The  experimental results showed that our k-nearest  policy is robust enough to outperform the baseline method even when we have five fewer agents in the test than the training.  

Second, K-nearest deep RL can address action dependency among highly interactive agents. This makes achieving optimal policy feasible in many industrial applications.   Third, by using weight-sharing, our proposed K-nearest deep RL algorithm   makes model management very simple as we only have to train and maintain one policy network for all of the groups. Unfortunately, simplicity  of the model training and maintenance  is not a high priority for many deep RL researchers. However, it is crucial in any industrial application. Finally, our experimental results showed that our solution converges even though  we used the experience replay.
This has a high practical value because the experience replay can improve the sample efficiency and reduce  the time and the computational costs  of the learning process.

\appendices
\section*{Appendix A}
\label{apendix}

Let say  score of agent $i$, $s_i$  is equal to the number of groups agent $i$  belongs to. Each agent at least belongs to one group, therefore:
 \begin{equation}
N \le \sum_{i=1}^N s_i,
\label{proof_Eq_1}
\end{equation}
Moreover, we know that each group has $k+1$ agents, therefore:
 \begin{equation}
\sum_{i=1}^N s_i = m\times(k+1).
\label{proof_Eq_2}
\end{equation}
Combining equations (\ref{proof_Eq_1}) and (\ref{proof_Eq_2}) we have:
 \begin{equation}
N \le m\times(k+1).
\label{proof_Eq_3}
\end{equation}
Since $m$ is an integer, we have:
 \begin{equation}
\ceil{\frac{N}{(k+ 1)}} \le m. 
\label{proof_Eq_3}
\end{equation}
Consider the case in which the groups do not share any agent. Since we have $k+1$   agents in each group, we have $m= \ceil{\frac{N}{(k+ 1)}}$. This shows that we cannot derive a lower bound higher than the one presented in equation (\ref{m_eq}). 

\begin{figure}[htbp]
\includegraphics[scale=.3]{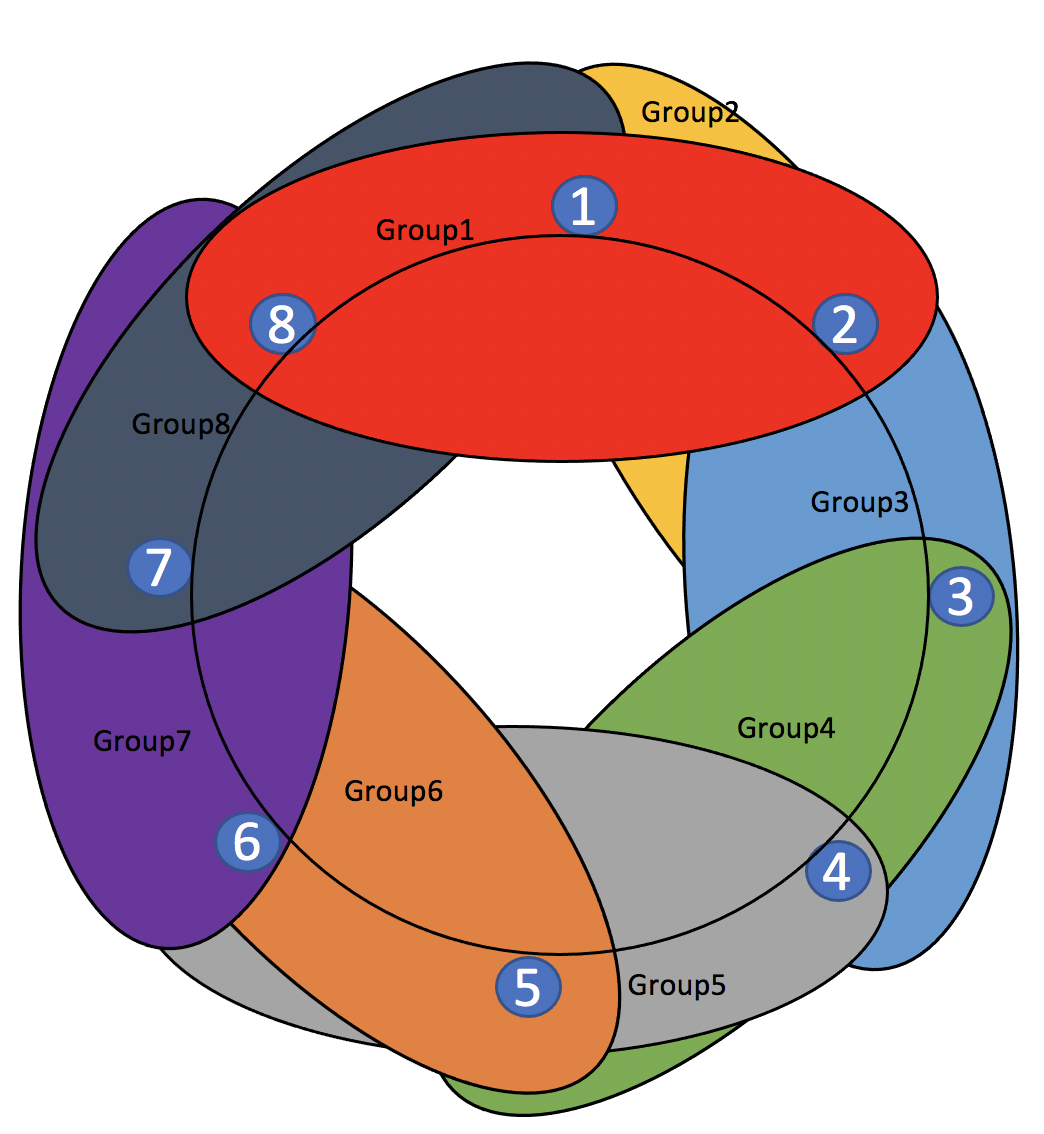}
\centering
\caption{ An example where number of groups is equal to the number of agents:  $N=8$ and $m=8$. }
\label{groupN}
\end{figure}

For the upper bound, it is trivial that we cannot have more than $N$ groups as the algorithm starts with assigning one group to each agent.
Therefore, we have:
 \begin{equation}
m \le N. 
\label{proof_Eq_3}
\end{equation}
Consider the case in which $N$ is greater than $3$,   $k=2$ and all the   agents are positioned in a circle   equally distanced from each other (see Figure \ref{groupN}).  In this case, each agents generates a group with the agent in each right and the agent in its left. This scenario shows a 
 rare example where we actually will end  up with $N$ groups. This shows that we cannot derive a  higher  bound higher lower  than the one presented in equation (\ref{m_eq}). 


\begin{thebibliography}{00}

\bibitem{Dota} C. Berner, G. Brockman, B. Chan, V. Cheung, P. Debiak, C. Dennison, D. Farhi, Q. Fischer,
S. Hashme, C. Hesse, et al., ``Dota 2 with large scale deep reinforcement learning,'' arXiv
preprint arXiv:1912.06680, 2019.
\bibitem{StarCraft} O. Vinyals, I. Babuschkin, J. Chung, M. Mathieu, M. Jaderberg, W. M. Czarnecki, A. Dudzik,
A. Huang, P. Georgiev, R. Powell, et al., ``Alphastar: Mastering the real-time strategy game
starcraft ii,'' DeepMind blog, p. 2, 2019.


\bibitem{Challenges} G. Dulac-Arnold, D. Mankowitz, and T. Hester, ``Challenges of real-world reinforcement learning,” arXiv preprint arXiv:1904.12901, 2019.

\bibitem{Challenges2} G. Dulac-Arnold, N. Levine, D. Mankowitz, J.  Li, C. Paduraru, S. Gowal, and T. Hester. ``An empirical investigation of the challenges of real-world reinforcement learning." arXiv preprint arXiv:2003.11881, 2020.
\bibitem{Challenges3} H. Khorasgani, H. Wang, and C. Gupta. ``Challenges of Applying Deep Reinforcement Learning in Dynamic Dispatching." arXiv preprint arXiv:2011.05570, 2020.
 \bibitem{variable quantity of agents} G. Wang and J. Shi, “Actor-critic for multi-agent system with variable quantity of agents,” in
International Conference on Internet of Things as a Service, pp. 48–56, Springer, 2018.
\bibitem{IDQ} V. Mnih, K. Kavukcuoglu, D. Silver, A. A. Rusu, J. Veness, M. G. Bellemare, A. Graves, M. Riedmiller, A. K. Fidjeland, G. Ostrovski, et al., ``Human-level control through deep reinforcement learning,” nature, vol. 518, no. 7540, pp. 529–533, 2015.

\bibitem{tsitsiklis1994asynchronous} J. N. Tsitsiklis, ``Asynchronous stochastic approximation and q-learning,” Machine learning,
vol. 16, no. 3, pp. 185–202, 1994

\bibitem{Multi-agent actor critic} R. Lowe, Y. I. Wu, A. Tamar, J. Harb, O. P. Abbeel, and I. Mordatch, ``Multi-agent actor critic for mixed cooperative-competitive environments,” in Advances in neural information
processing systems, pp. 6379–6390, 2017.
\bibitem{Value-decomposition} P. Sunehag, G. Lever, A. Gruslys, W. M. Czarnecki, V. F. Zambaldi, M. Jaderberg, M. Lanctot,
N. Sonnerat, J. Z. Leibo, K. Tuyls, et al., ``Value-decomposition networks for cooperative
multi-agent learning based on team reward.,” in AAMAS, pp. 2085–2087, 2018.
\bibitem{Qmix} T. Rashid, M. Samvelyan, C. S. De Witt, G. Farquhar, J. Foerster, and S. Whiteson, ``Qmix:
Monotonic value function factorisation for deep multi-agent reinforcement learning,” arXiv
preprint arXiv:1803.11485, 2018.
\bibitem{Foerster} J. Foerster, I. A. Assael, N. De Freitas, and S. Whiteson, `` Learning to communicate with deep
multi-agent reinforcement learning,” in Advances in neural information processing systems,
pp. 2137–2145, 2016.
\bibitem{HitachiDynamicDispatching} C. Zhang, P. Odonkor, S. Zheng, H. Khorasgani, S. Serita, C. Gupta, and H. Wang. ``Dynamic dispatching for large-scale heterogeneous fleet via multi-agent deep reinforcement learning." In 2020 IEEE International Conference on Big Data (Big Data), pp. 1436-1441. IEEE, 2020.
\bibitem{SAC} T. Haarnoja, A. Zhou, P. Abbeel, and S. Levine. "Soft actor-critic: Off-policy maximum entropy deep reinforcement learning with a stochastic actor." In International conference on machine learning, pp. 1861-1870. PMLR, 2018.
\end{thebibliography}
\end{document}